\documentclass[doublecolumn]{IEEEtran}
\usepackage{amsmath,amsthm,amsfonts,amssymb,bm,mathrsfs}
\usepackage{subfigure}
\usepackage{graphicx,multicol}
\usepackage{algorithm}
\usepackage{algorithmic}
\usepackage{multirow,booktabs,threeparttable}
\usepackage{color,cite,url}
\usepackage{booktabs}
\usepackage{amssymb}
\usepackage{caption}
\usepackage{tabularx}
\usepackage{makecell}

\newcolumntype{Y}{>{\centering\arraybackslash}X}
\newcolumntype{L}{>{\raggedright\arraybackslash}X}

\graphicspath{{fig/}}

\usepackage{setspace}
\theoremstyle{remark}

\theoremstyle{plain}

\begin{document}
\title{NetGPT: An AI-Native Network Architecture for Provisioning Beyond Personalized Generative Services}

 \author{
 	Yuxuan Chen, Rongpeng Li, Zhifeng Zhao, Chenghui Peng, Jianjun Wu, Ekram Hossain, and Honggang Zhang

 	\thanks{Y. Chen and R. Li are with Zhejiang University, Hangzhou 310027, China, (email: \{cyx00, lirongpeng\}@zju.edu.cn).}

 	\thanks{C. Peng and J. Wu are with Huawei Technologies Co., Ltd., Shanghai 210026, China (email: \{pengchenghui,wujianjun\}@huawei.com).}

	\thanks{Z. Zhao and H. Zhang are with Zhejiang Lab, Hangzhou 310012, China as well as Zhejiang University, Hangzhou 310027, China (email: \{zhaozf,honggangzhang\}@zhejianglab.com).}

    \thanks{E. Hossain is with University of Manitoba, Winnipeg, Manitoba, Canada (email: ekram.hossain@umanitoba.ca).}
}

\maketitle

\begin{abstract}
	Large language models (LLMs) have triggered tremendous success to empower our daily life by generative information. The personalization of LLMs could further contribute to their applications due to better alignment with human intents. Towards personalized generative services, a collaborative cloud-edge methodology is promising, as it facilitates the effective orchestration of heterogeneous distributed communication and computing resources. In this article, we put forward \texttt{NetGPT} to capably synergize appropriate LLMs at the edge and the cloud based on their computing capacity. In addition, edge LLMs could efficiently leverage location-based information for personalized prompt completion, thus benefiting the interaction with the cloud LLM. In particular, we present the feasibility of \texttt{NetGPT} by leveraging low-rank adaptation-based fine-tuning of open-source LLMs (i.e., GPT-2-base model and LLaMA model), and conduct comprehensive numerical comparisons with alternative cloud-edge collaboration or cloud-only techniques, so as to demonstrate the superiority of \texttt{NetGPT}. Subsequently, we highlight the essential changes required for an artificial intelligence (AI)-native network architecture towards \texttt{NetGPT}, with emphasis on deeper integration of communications and computing resources and careful calibration of logical AI workflow. Furthermore, we demonstrate several benefits of \texttt{NetGPT}, which come as by-products, as the edge LLMs' capability to predict trends and infer intents promises a unified solution for intelligent network management \& orchestration. We argue that \texttt{NetGPT} is a promising AI-native network architecture for provisioning beyond personalized generative services.
\end{abstract}

\section{Introduction}
With the remarkable success of deep learning spanning from decision-making in AlphaGo to human-level interaction like ChatGPT, it is anticipated that artificial intelligence (AI) will be embodied in 6G networks. Along with the enhanced edge computing capabilities, AI could benefit the effective orchestration of network resources and improve the quality of service (QoS). Correspondingly, investigation on efficient AI-based service provisioning has attracted intense research interest. On the other hand, the application of one AI model is often limited to certain scenarios or tasks. 
In this context, large language models (LLMs) (e.g., generative pre-trained transformer, GPT) could perform well in various natural language processing (NLP) and computer vision tasks. These inspiring advancements shed light on revolutionizing cellular networks by LLMs. For instance, \cite{zou2023wireless} harnesses collective intelligence for efficient network management, by delving into the deployment of on-device LLMs and proposing a multi-agent system architecture. Similarly, \cite{wang2023build} challenges traditional network paradigms for LLM training, and proposes a novel, cost-effective architecture tailored to LLM-specific communication patterns, with a demonstrated $75\%$ network cost reduction without sacrificing the performance. These progresses \cite{zou2023wireless,wang2023build} underscore the importance of integrating LLMs with innovative network architectures, a key to unlock greater efficiency and performance in advanced network environments. Notably, towards provisioning personalized generative services (e.g., personalized assistance and recommendation systems), fine-tuning is still a prerequisite to align pre-trained LLMs to follow human intents \cite{zhang_building_2023} and yield personalized outputs. Nevertheless, it might be cost-ineffective to simply deploy multiple copies of bloated model parameters to support different purposes, and a feasible solution remains under-investigated.

In order to boost the personalization of LLMs, a collaborative cloud-edge methodology is essential \cite{xu_unleashing_2023}. Compared to the cloud-only LLM deployment, such a cloud-edge collaboration enjoys multi-folded merits. Firstly, it provides more freedom to allow edge servers to deploy various fine-tuned LLMs and adapt to environmental differences, thus making the service personalization and customization possible. Meanwhile, it contributes to bridging data-abundant generative devices with more adjacent servers. Therefore, it could reduce the latency and save the communication overhead to upload all data to more remote cloud servers. Incorporating generative LLMs into the edge networks promises to facilitate the effective utilization of communication and computing (C\&C) resources.

\begin{figure*}[ht]
	\centering
	\includegraphics[width=0.85\textwidth]{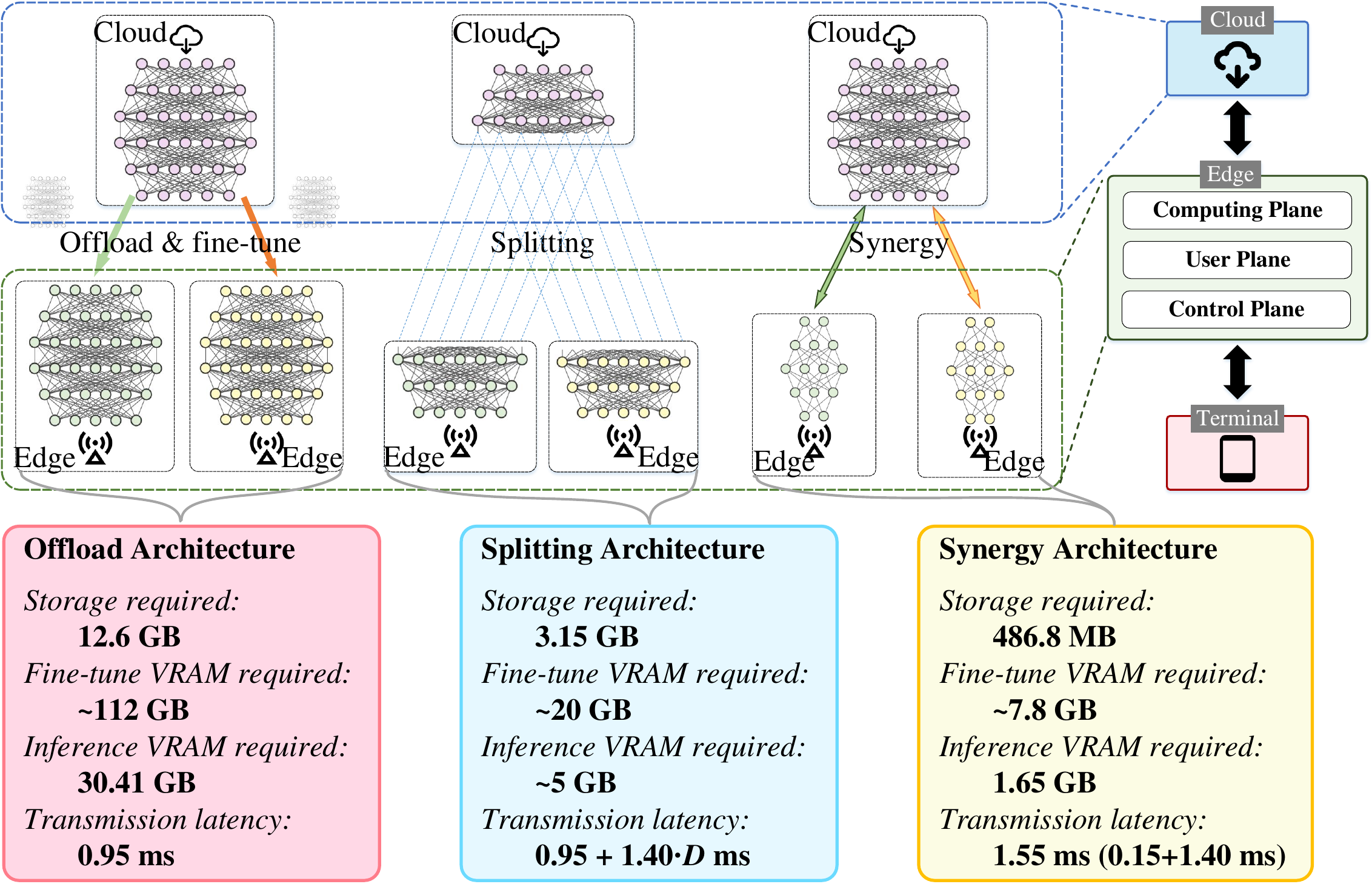}
	\caption{An illustration of candidate means to realize the could-edge collaboration for \texttt{NetGPT} and with comparison from alternative cloud-edge frameworks. Specifically, transmission latency is calculated for $10,000$ ``concise prompts'' with an average size of $12$ bytes (correspondingly $95$-byte ``comprehensive prompt'') and a transmission rate of $1$ Gbps. For the ``LLM Splitting'' framework, we take an example of splitting $1/4$ of the LLaMA-7B model at the edge, with $D \approx 10,922$ representing the ratio of intermediate layer data volume to input token size. }
	\label{fig:options}
\end{figure*}

As illustrated in Fig. \ref{fig:options}, there are several distinctive ways to implement the cloud-edge collaboration for deployment of LLMs (e.g., local fine-tuning, model splitting). Specifically, by \emph{offloading} cloud-trained LLMs, local edge servers tailor the cloud-trained LLMs to accommodate personalized and customized services based on the user preference and specified scenarios. 
However, such an approach might face severe implementation issues in practice, as repetitive fine-tuning of complete LLMs implies significant computational burden, and also distributed deployment of proprietary LLMs might raise intellectual property concerns from model developers. Meanwhile, force-fitting an entire LLM on edge possibly strains the limited computing resources of edge servers and makes the cost of edge computing unacceptable.
Alternatively, \emph{splitting} LLMs to cloud and edge servers \cite{wu_split_2023}, by deploying some layers of large-scale deep neural network (DNNs) at the edge while leaving the remaining layers to the cloud, can effectively balance the disproportionate computing resources of edge and cloud servers. Within the model splitting, how to effectively partition the DNNs between the edge and the cloud belongs to one of the most challenging issues, as it should minimize the end-to-end latency while maintaining a sufficiently small model size for the edge servers \cite{wu_split_2023}. Such a model partitioning can be even more intricate, given billions of parameters in a typical LLM. 
Besides, the LLMs might leak private details from the data for training \cite{kandpal_deduplicating_2022}. In other words, it might be challenging to directly adopt both local fine-tuning and model splitting as an implementation means of collaborative cloud-edge methodology.

In this article, we put forward \texttt{NetGPT} that aims to respect the cloud-edge resource imbalance and synergize different sizes of functional LLMs at the edge and cloud, thus promising to foster improved prompt responses and personalized outputs. Specifically, in apparent contrast to AI-exogenous network with decoupled C\&C resources, \texttt{NetGPT} could leverage converged C\&C to deploy smaller edge LLMs for the edge while larger one for the cloud, and meaningfully realize collaborative cloud-edge computing to provision personalized content generation services. Besides, \texttt{NetGPT} incorporates a logical AI workflow that could be developed to determine performance-consistent communication links. For example, in \texttt{NetGPT}, the performance-driven communication link could terminate at the edge to accelerate the response assuming the availability of satisfactory edge LLM-induced content. Otherwise, inspired by the idea of prompt learning \cite{ liu_pretrain_2023}, the LLMs at the edge can infer the context and actively append (or fill in) some local or personalized information, so as to acquire a more comprehensive result at the cloud. Furthermore, as a by-product, the edge LLMs contribute to a unified solution for intelligent network management \& orchestration (e.g., user intent inference and popularity prediction). Therefore, consistent with the trend to deeply integrate C\&C, \texttt{NetGPT} represents an AI-native LLM synergy architecture and implies the enhanced collaboration between edge and cloud LLMs.

\section{Implementation Showcase of NetGPT}
\label{sec:implementation}
\begin{figure*}[t]
	\centering
	\includegraphics[width=0.975\textwidth]{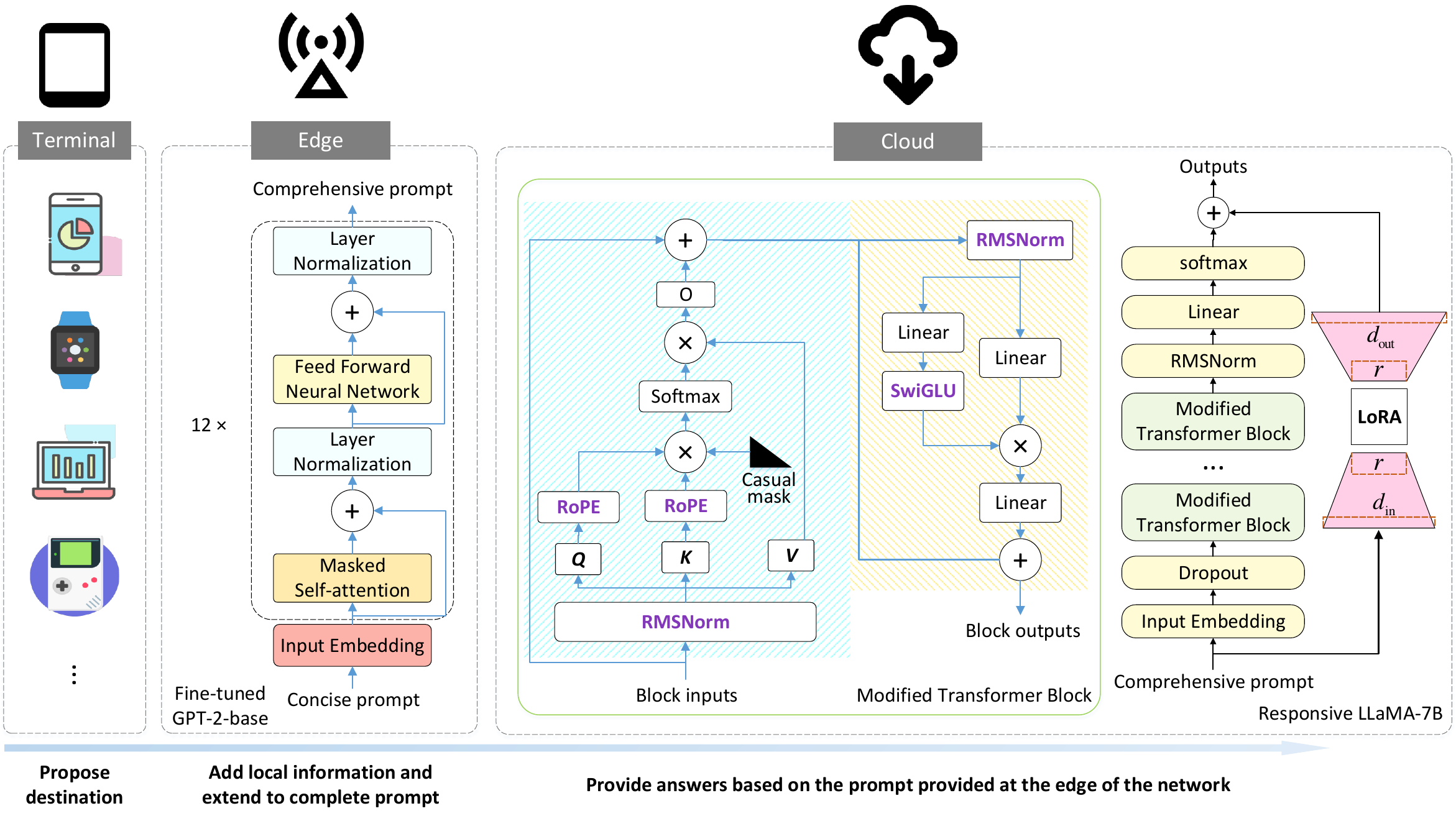}	\caption{A framework of collaborative cloud-edge computing towards \texttt{NetGPT}.}
	\label{fig:orchestration}
\end{figure*}

As illustrated in Fig. \ref{fig:orchestration}, we present a synergistic cloud-edge framework to accomplish personalized generative services, by leveraging distinctive pre-trained LLMs for cloud and edge (e.g., base stations [BSs]) deployment. In particular, limited by the availability of open-source LLMs, we select and deploy the LLaMA-7B model \cite{touvron_llama_2023} and the GPT-2-base model, which consist of approximately $6.7$ and $0.1$ billion parameters, at the cloud and the edge, respectively. However, it should be noted that \texttt{NetGPT} allows the utilization of other LLMs as well. On this basis, we delve into implementation details of cloud-edge LLM synergy towards \texttt{NetGPT} in an incremental manner. In particular, we start with detailed DNN structures of two LLMs (i.e., LLaMA-7B model and GPT-2-base model). Then, we discuss the effective means to fine-tune these LLMs on computation-limited devices, and demonstrate the effectiveness of synergizing edge LLMs and cloud LLM for location-based personalized generative services. Notably, the ``LLM synergy'' framework significantly contrasts with split learning \cite{wu_split_2023} and federated learning \cite{zhang_building_2023}, which aims to train the divided segments of a DNN on different clients, or jointly learn from data distributed across multiple nodes in a data privacy-friendly manner. Orthogonal to split learning and federated learning, our framework focuses on effective cloud-edge collaboration with prompt enhancement \& de-duplication at the edge and personalized responses at the cloud.

\subsection{DNN Structure of LLMs at the Edge and Cloud}
\subsubsection{DNN structure of GPT-2-base model:}
The GPT-2-base model, which is the smallest version of the GPT-2 series, encompasses $12$ stacked layers of the original transformer structure (i.e., an $8$-head self-attention sublayer and an FNN sublayer). A fixed absolute positional encoding of sine and cosine positions is employed to pre-transform the input sequence. In addition, GPT-2 leverages a rectified linear unit (ReLU) activation function (i.e., $f_{\text{ReLU}}(x) = \max(0,x)$). Due to its relatively exceptional performance and minimal computational requirements, it can be appropriate to be deployed at the network edge.

\subsubsection{DNN structure of LLaMA model:}
LLaMA, which is trained on a large set of unlabeled data and is ideal for fine-tuning for downstream tasks, features various parameter versions as well \cite{touvron_llama_2023}. Compared to GPT-3, LLaMA incorporates several specific enhancements to maintain similar performance while significantly reducing the number of parameters~\cite{touvron_llama_2023}. For example, in order to enhance training stability, LLaMA normalizes the input of each sub-layer instead of normalizing the output. Moreover, it adopts the root mean square layer normalization (RMSNorm) function as a simplified replacement for layer normalization, by employing the root mean square (RMS) rather than the standard deviation. Additionally, RMSNorm introduces a learnable scaling factor that enables adaptive feature scaling. Thus, it contributes to improving normalization effects across diverse features with distinctive value ranges. Secondly, LLaMA replaces the ReLU activation function with Swish-gated linear unit (SwiGLU) \cite{shazeer_glu_2020}, which combines the Swish function (i.e., $f_{\text{Swish}}(x)=x\cdot \sigma (\beta x)$ with $\sigma (x) = \frac{1}{1+e^{- x}}$ and a trainable parameter $\beta$) with GLU (i.e., $f_{\text{GLU}}(x) = x \cdot \sigma(Wx + b)$ parameterized by trainable parameters $W$ and $b$), thereby possibly activating neurons according to the input in a more selective manner and imparting smoothness to effectively capture intricate non-linear relationships. Lastly, LLaMA introduces rotary position embedding (RoPE)~\cite{su_roformer_2022}, which encodes positional information with a pre-defined rotation matrix and naturally incorporates explicit relative position dependency in the self-attention formulation. Compared to absolute position encoding which assigns a distinct encoded representation to each position in the sequence, the taken form of relative position encoding in RoPE enables a more effective modeling of long-range dependencies within the contextual information. Thereby, RoPE could align with intuitive understanding and exhibits superior performance in practice.

\subsection{Low-Rank Adaptation and Cloud LLM Fine-Tuning}
As LLaMA lacks the capability to generate responsive text \cite{touvron_llama_2023}, an extra fine-tuning is still required. However, a direct fine-tuning of LLMs such as a LLaMA still requires significant computational resources. For example, it demands $112$ GB video random access memory (VRAM) to fine-tune the LLaMA-7B model, far more than the capacity of NVIDIA A100 Tensor Core GPU. Therefore, we leverage a low-rank adaptation (LoRA) technique~\cite{hu_lora_2022}, a parameter efficient fine-tuning (PEFT) technique, to achieve parameter-efficient fine-tuning on a consumer-level hardware. Notably, for \texttt{NetGPT}, PEFT techniques are not indispensable components. Instead, LoRA just demonstrates the feasibility of fine-tuning on computation-limited network elements of \texttt{NetGPT}.

In particular, in order to fine-tune a complete parameter matrix $\bm{W} \in \mathbb{R}^{d_{\text{in}}\times d_{\text{out}}}$, LoRA specially adds a bypass pathway to simulate the matrix update $\Delta \bm{W}$ by using two downstream parameter matrices $\bm{A} \in \mathbb{R}^{d_{\text{in}}\times r}$ and $\bm{B} \in \mathbb{R}^{r\times d_{\text{out}}}$ with the intrinsic rank $r$. In other words, under the condition that $r \ll \min(d_{\text{in}},d_{\text{out}})$, LoRA successfully transforms large parameter matrix $\Delta \bm{W}$ into lower-rank dimensional ones with $\Delta \bm{W} \approx \bm{A}\bm{B}$. Our experiment shows that it only costs $28$ GB VRAM to fine-tune the LLaMA-7B model, without significantly elongating the training duration. Additionally, the required storage space for fine-tuning could be greatly reduced from $12.55$ GB to $50$ MB\footnote{Such statistics are obtained under the configuration that $r=8$ and a learning rate-related scalar factor equals $16$.}. On the basis of LoRA, we can utilize the Stanford Alpaca dataset \cite{alpaca} to fine-tune LLaMA-7B model and obtain a qualified responsive LLaMA-7B model. 

\subsection{Edge LLM Fine-Tuning}
\subsubsection{Mathematical formulation:}
In the ``LLM synergy'' framework, the edge node plays a pivotal role in prompt enhancement. This transformation from ``concise prompt'' $P_{\text{con}}$ to ``comprehensive prompt'' $P_{\text{com}}$ can be mathematically formulated as 
$P_{\text{com}} = \text{LLM}_{\theta}(P_{\text{con}}; \mathcal{I}_{\text{personalized}})$, 
where $\text{LLM}_{\theta}$ represents the processing function of the edge LLM parameterized by $\theta$, and $\mathcal{I}_{\text{personalized}}$ encompasses localized or personalized information. Notably, the transformation process leverages the astonishing generative capability of LLM in a non-transparent ``black box'' manner. Meanwhile, the transformation effectiveness has been validated in Fig. \ref{fig:sample}.\\ \indent On the other hand, this transformation requires a fine-tuning process of the edge LLM on collected dataset $\mathcal{D}$, which can be conceptualized as an optimization problem. In other words, it is equivalent to finding an appropriate set of parameters $\theta$ that could minimize the cumulative loss between edge LLM-generated ``comprehensive prompts'' and human-intended ones $P_{\text{intent}}$. Mathematically,
    $\theta^* = \underset{\theta}{\mathrm{arg\, min}} \sum_{(P_{\text{con}}, P_{\text{intent}}) \in \mathcal{D}} \mathbb{L}(\text{LLM}_{\theta}(P_{\text{con}}; \mathcal{I}_{\text{personalized}}), P_{\text{intent}})$,
where $\mathbb{L}$ denotes a loss function (e.g., cross-entropy).
\subsubsection{LLM-instructed data collection:}
Different from the cloud LLM, which is fine-tuned with web-based conversational datasets \cite{alpaca}, so to respond to ``comprehensive prompts'' in a manner that is informed, contextually rich, and aligns with the broader conversational context, the data collection for edge LLMs requires extra efforts. This differentiation in data sources underlines the collaborative efficacy of the synergy architecture, with each component playing a specialized role in the data processing workflow.
In order to implement personalized edge LLM, it is crucial to grant the GPT-2-base model the capability to extend a ``concise'' prompt by appending location-based information. Basically, the positioning information can be conveniently obtained according to the locations of affiliated BSs stored in the 5G access and mobility management function (AMF). Meanwhile, in order to complement more comprehensive information, we take the self-instruct approach~\cite{selfinstruct} and leverage OpenAI's Text-Davinci-003 model to generate useful text samples. In particular, as for each location, we use a set of manually written location-related prompts to interact with the OpenAI's Text-Davinci-003 model, and leverage the generative response texts as the ``intended prompt'', which augments the most likely word following the ``concise prompt'' and describes the corresponding more comprehensive intent. For example, as illustrated on the top-left side of Fig. \ref{fig:sample}, from a perspective of real-life linguistic patterns, ``libraries'' frequently correlates with the contextual word ``collections''.
Besides, the top-right part of Fig. \ref{fig:sample} demonstrates how post-fine-tuning edge LLM can generate an elongated and more comprehensive prompt in response to the ``concise prompt''. On this basis, a series of mappings between the ``concise prompt'' and an ``intended prompt'' can be collected. Considering the size and task complexity of the edge LLM, we collect a dataset comprising approximately $4,000$ samples for directly fine-tuning the GPT-2-base model towards a prompt-completion model. Notably, for scenarios where stronger generality is required, edge LLMs can be enhanced with a larger-scale LLM, and fine-tuning techniques such as LoRA can be employed as well.

\subsection{Performance Showcase}

\begin{figure*}[t]
	\centering
	\includegraphics[width=0.975\textwidth]{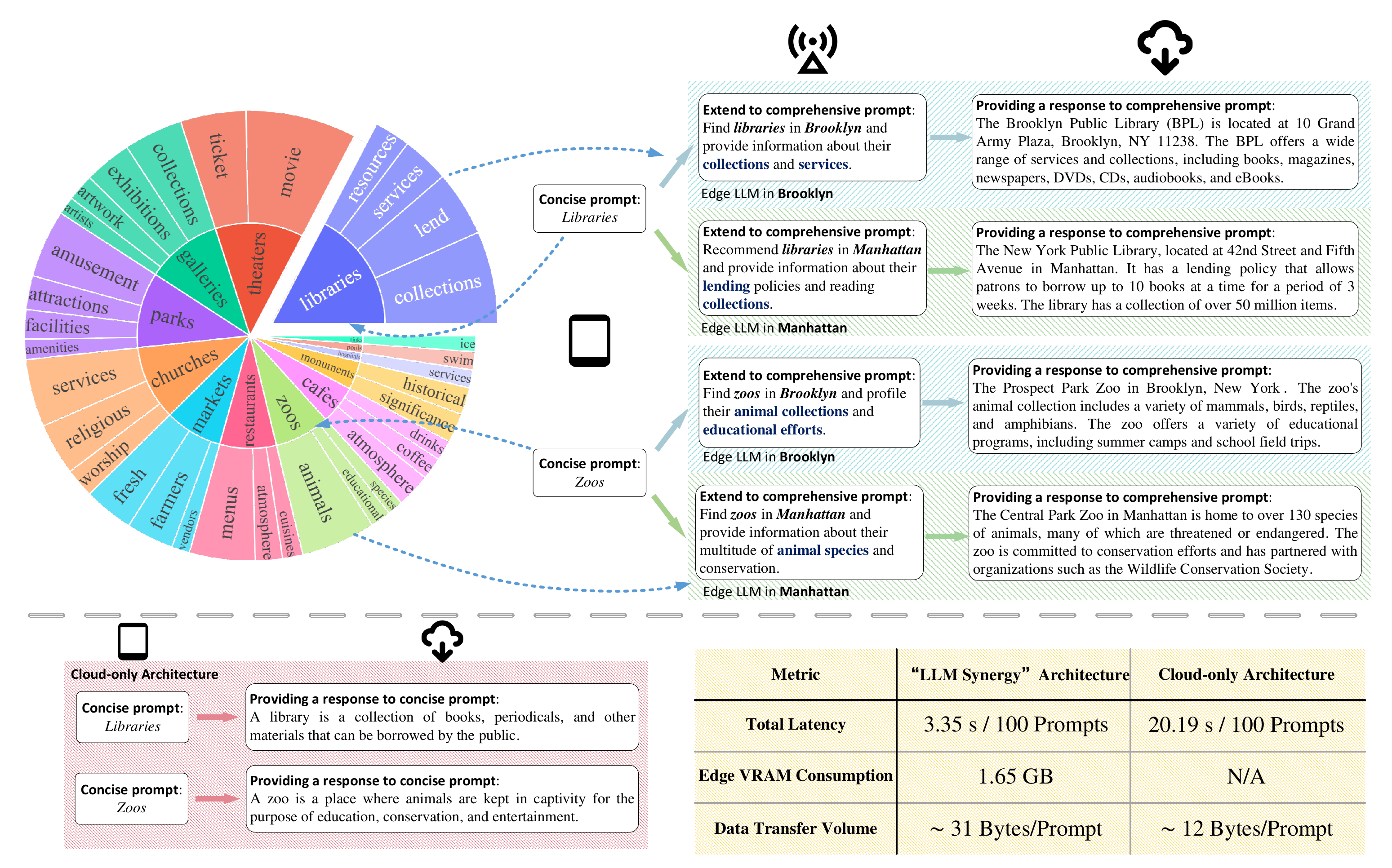}
	\caption{Comparison between ``LLM synergy'' framework and cloud-only solution. Top-left: Inferring contextual words following ``concise prompts''. Top-right: Examples of generated ``comprehensive prompts'' by regional edge LLMs under ``LLM synergy'' framework, as well as more personalized cloud LLM responses. Bottom-left: Simpler, non-personalized responses from cloud-only solution for the same prompts. Bottom-right: Numerical comparison between ``LLM synergy'' and cloud-only frameworks.}
	\label{fig:sample}
\end{figure*}
Fig. \ref{fig:sample} further demonstrates the performance of \texttt{NetGPT}. In particular, as illustrated in Fig. \ref{fig:sample}, the edge LLM is capable of complementing the ``concise prompt'' according to the chart at the top-left part of Fig. \ref{fig:sample}, which highlights most frequently used words for generating each corresponding ``comprehensive prompt''. Furthermore, different BSs add location-based personalized information so as to satisfy distinctive requirements. Subsequently, the edge LLM processes the user-submitted ``concise prompt'', and feeds the complemented prompt to the cloud. Next, a more complete generative response could be anticipated. It can be observed from the top-right part of Fig. \ref{fig:sample} that \texttt{NetGPT} could yield different location-based responses, which manifests the capability to handle personalized generative services through effective cloud-edge synergy.

The LLM synergy brings multiple merits (e.g., context-aware processing and prompt optimization) within a resource-constrained environment. In particular, the edge LLM lays the groundwork for efficient cloud processing through personalized treatment and request optimization. In other words, the ``LLM synergy'' framework competently processes multiple requests at the edge in a batch manner. Additionally, the ``LLM synergy'' framework could take advantage of the de-duplication capability of the edge to filter redundant requests. Therefore, it promises to further reduce the communication burden, and is more qualified for prompt-intensive application scenarios. On the other hand, the ``LLM synergy'' framework strengthens data privacy and security, as the data processing capability of edge nodes also promises to limit the transmission of sensitive data and reduce potential security risks. Meanwhile, the high-performance processing capabilities of the cloud LLM ensure the quality and complexity of request handling, thus collectively facilitating an efficient and accurate end-to-end service provisioning.
\subsection{Discussions}

\subsubsection{Comparison with cloud-edge solutions:}

As shown in Fig. \ref{fig:options}, we compare the three cloud-edge collaborative frameworks in terms of the requirements on storage, fine-tuning and inference VRAM, as well as the end-to-end transmission latency. Notably, in the experiments, ``concise prompts'' consume averaging $12$ bytes, while ``comprehensive prompts'' correspond to $95$ byte consumption on average. On this basis, the transmission latency is computed by processing $10,000$ at a transmission rate of $1$ Gbps for both the end-to-edge and edge-to-cloud links.
Besides, for the ``LLM splitting'' framework, how to determine an ideal point for model division, which appropriately balances communication cost and performance, remains a critical challenge. Therefore, we showcase the results after partitioning $1/4$ of the LLaMA-7B model at the edge. In this case, if we use $D$ to denote the quantified ratio between data volume in the model's middle layer and the input token size, we have
$
    D := \frac{\text{hidden layer dimension} \times \text{data type size}}{\text{average input size}} = \frac{4096 \times 4}{12} \approx 10,922
$.

By Fig. \ref{fig:options}, the ``LLM offload'' framework can directly handle the prompts dependent on the offloaded LLM at the edge, and thus could more significantly reduce the transmission delay. Nevertheless, such a benefit comes at the expense of significant storage and computing resource consumption. Meanwhile, the ``LLM splitting'' framework could alleviate the resource requirements on the edges, but suffer from the engineering challenge of determining an appropriate splitting point without sacrificing model performance. Furthermore, ``LLM splitting'' inevitably adds communication overhead. On the contrary, the ``LLM synergy'' framework demands minimal storage and computing resources at the edge without sacrificing the essential efficiency, which manifests its superiority in terms of both performance and flexibility.

\subsubsection{Comparison with cloud-only solution:}

In the aforementioned scenario, in order to simultaneously transmit and process $100$ ``concise prompts'' averaging $12$ bytes under a transmission rate of $1$ Gbps, the ``LLM synergy'' framework yields an end-to-end latency of just $3.35$ seconds and significantly outperforms the cloud-only solution, which instead requires $20.19$ seconds. This performance superiority lies in that in cloud-only setups, the inherent queuing latency gets exacerbated by high volumes of concurrent requests and each individual request necessitates the establishment of an independent communication connection with the cloud infrastructure, potentially leading to increased handshake signaling overheads and more frequent re-transmissions.

Moreover, as opposed to the cloud-only approach's lack of edge resource usage, the ``LLM synergy'' framework consumes approximately $1.65$ GB edge VRAM for obtaining ``comprehensive prompts''. Through the enhancement of edge computing, the cloud in ``LLM synergy'' framework places a personalized prompt generator on the edge to improve the resource efficiency, so that the cloud can generate personalized content without increasing the computational cost. Meanwhile, for each request, ``LLM synergy'' framework transfers approximately $31$ bytes, compared to the cloud-only approach's around $12$ bytes. Despite the higher data transfer volume, for the bandwidth-abundant edge-cloud links, the extra tokens in the ``comprehensive prompt'' sound trivial. Also, the ``LLM synergy'' framework ensures that the communication is more meaningful by providing a richer context for the cloud's language model, leading to more personalized and accurate output content. Given identical concise prompt inputs, as demonstrated in Fig. \ref{fig:sample}, the ``LLM synergy'' framework is able to generate more specific and personalized output content. In summary, the ``LLM synergy''-based \texttt{NetGPT} exhibits superior performance over cloud-only solution.

\subsubsection{Extension to large multi-modal models (LMMs)}

The emergence of LMMs marks a significant leap beyond text-only tasks and \texttt{NetGPT} could naturally benefit from the integration with LMMs. For example, deploying smaller, versatile LMMs at the edge allows for the collection and refinement of local data, and generates prompts that are more comprehensively aligned with immediate surroundings. On the other hand, powerful LMMs located in the cloud, such as DALL·E and Sora, contribute to generating detailed multi-modal responses on top of enriched information from the edge. Notably, the effective integration with LMMs requires further attention due to multi-modal dataset scarcity, increased model complexity, and semantic gap across modalities.
\section{AI-Native Network Architecture Towards NetGPT}
\label{sec:architecture}
\begin{figure*}
	\centering
	\includegraphics[width=0.975\textwidth]{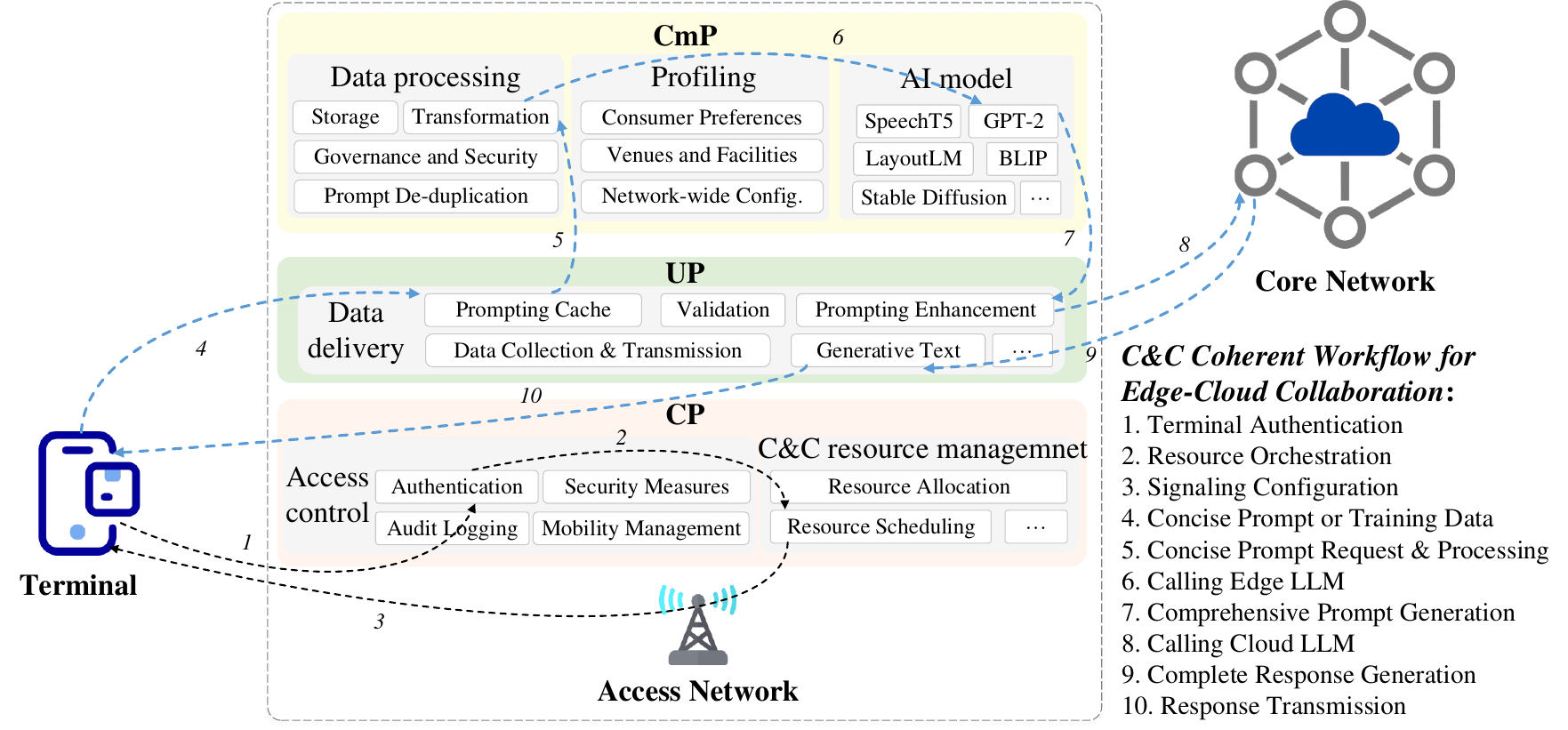}
	\caption{The illustration of an AI-native network architecture and logical AI workflow for \texttt{NetGPT}.}
	\label{fig:workflow}
\end{figure*}

We argue that \texttt{NetGPT} provides the opportunity to transform cellular networks into an AI-native networking architecture, which provisions personalized, networked and inclusive intelligence for end users and grants users more privilege to access generative services anytime and anywhere. Nevertheless, such a transformation does come at a cost. It requires substantial changes, far more than installing racks of servers at the edge location and local break-out of the traffic for edge processing. In particular, compared with conventional connectivity-oriented communications systems, wherein a typical service establishes connections between two specific terminals and the communication source and destination are clearly defined by end users, \texttt{NetGPT} requires to establish generative performance-driven connections more implicitly. Moreover, as \texttt{NetGPT} involves more frequent data collection and processing modules for training personalized LLM models, computing resources shall be consistently scheduled to accomplish \texttt{NetGPT}, and an independent computing plane (CmP), which coordinates computing resources and perform AI-related functionalities, becomes complementary to the control plane (CP) and user plane (UP). In other words, as shown in Fig. \ref{fig:workflow}, \texttt{NetGPT} necessitates the design of deeply converged C\&C in radio access networks (RANs). On top of these novel features, a logical AI workflow shall be devised to establish (beyond) generative service orchestration.


\subsection{Converged C\&C Resource Management}
As part of provisioned services in future cellular networks, the orchestration of resources for \texttt{NetGPT} shares some similarities with that for other network services, including seamless connectivity control in the CP and reliable information transmission in the UP. However, it also poses additional challenges. For example, for the advocated CmP in \texttt{NetGPT}, orchestrating heterogeneous computing resources effectively is of paramount importance, considering that the central processing units (CPUs) capably handle general computing tasks, while the graphics processing units (GPUs) specialize in parallel processing tasks such as neural network computations, and the video processing units (VPUs) excel in the real-time image and video processing. Therefore, it is crucial to accommodate tasks with the most appropriate resources to enhance system efficiency and performance, in an orchestrated manner with intelligent adaptation. In addition, since the scope of resources spans distributed nodes from the cloud to the terminal, novel protocol stack needs be carried on radio control signaling (e.g., RRC) or radio data protocol (e.g. PDCP or SDAP) to transmit AI-generative messages and implement model update \& distribution.
In that regard, introducing new RRC messages or flags that would prioritize AI data, could ensure rapid allocation of network resources for real-time AI workflows. Furthermore, as quantization techniques are widely used in developing light-weight AI models, refining PDCP header compression algorithms to better identify and compress AI data payloads promises improved data processing efficiency. Additionally, defining a dynamic configuration module/sublayer within the protocol stack, tailored to specific AI tasks, allows for the establishment of a closed-loop mechanism for real-time adjustments.

\subsection{Data Processing and Privacy Protection}
As discussed in Section \ref{sec:implementation}, data processing (e.g., data collection and fine-tuning) is heavily leveraged to lay the very foundation for producing generative LLM models. Besides collecting and storing data, it is feasible to filter duplicate prompts at the edge, so as to reduce the communication burden.
In addition, it is essential to introduce data desensitization modules as key data processing services, so as to avoid privacy risks and protect the privacy embedded in the data. Meanwhile, data policy enforcement modules, which handle data according to regulatory as well as non-regulatory rules (e.g., geographic restrictions), will be executed by default to ensure the integrity and legitimation of data processing. Moreover, contingent on the regulation and data usage policy, it is also feasible to devise some data processing model libraries and expose the capabilities with appropriate access control for entities to utilize the data services.

\subsection{Personalized Profiling}
In order to create a highly customized \texttt{NetGPT}, location-oriented profiling shall be significantly enhanced to support the definition and operation of personalized generative AI services. For example, local venue and facility information can be specially gathered to train edge LLMs. On the other hand, user service nodes (USN) can contain customized services at end-user level as well, so as to meet diversified customer requirements. Meanwhile, it could further support to establish the user profiling and characterize connected terminals.

\subsection{Logical AI Workflow}

In order to effectively provision AI services, it is critical to develop some logical AI workflows to parse and orchestrate \texttt{NetGPT} services. Notably, a logical AI workflow, which facilitates a set of network functions physically distributed at both the edge and the cloud to coherently deliver ``concise prompt'', ``comprehensive prompt'' and ``generative responses'', regulates data processing and profiling to train personalized LLMs at the CmP. Furthermore, logical AI workflows are mapped to physical resources during service deployment, so as to take into account the QoS requirements of related services. Notably, as the workflow covers a wide scope of network functions, the processing may be serial or directed acyclic graph-based, and thus involves comprehensive optimization techniques beyond the scope of this article. On the other hand, the logical AI workflow is not limited to generative services. As discussed in Section \ref{sec:benefit} lately, the logical AI workflow significantly contributes to the improvement of QoS in a more customizable manner.

\section{LLM-Based Unified Solution for Network Management \& Orchestration}
\label{sec:benefit}
Apart from providing personalized generative services, \texttt{NetGPT} and the AI-native architecture could provide a unified solution for intelligent network management \& orchestration, on top of deployed edge LLMs. 

\begin{figure*}[t]
	\centering
	\includegraphics[width=0.95\textwidth]{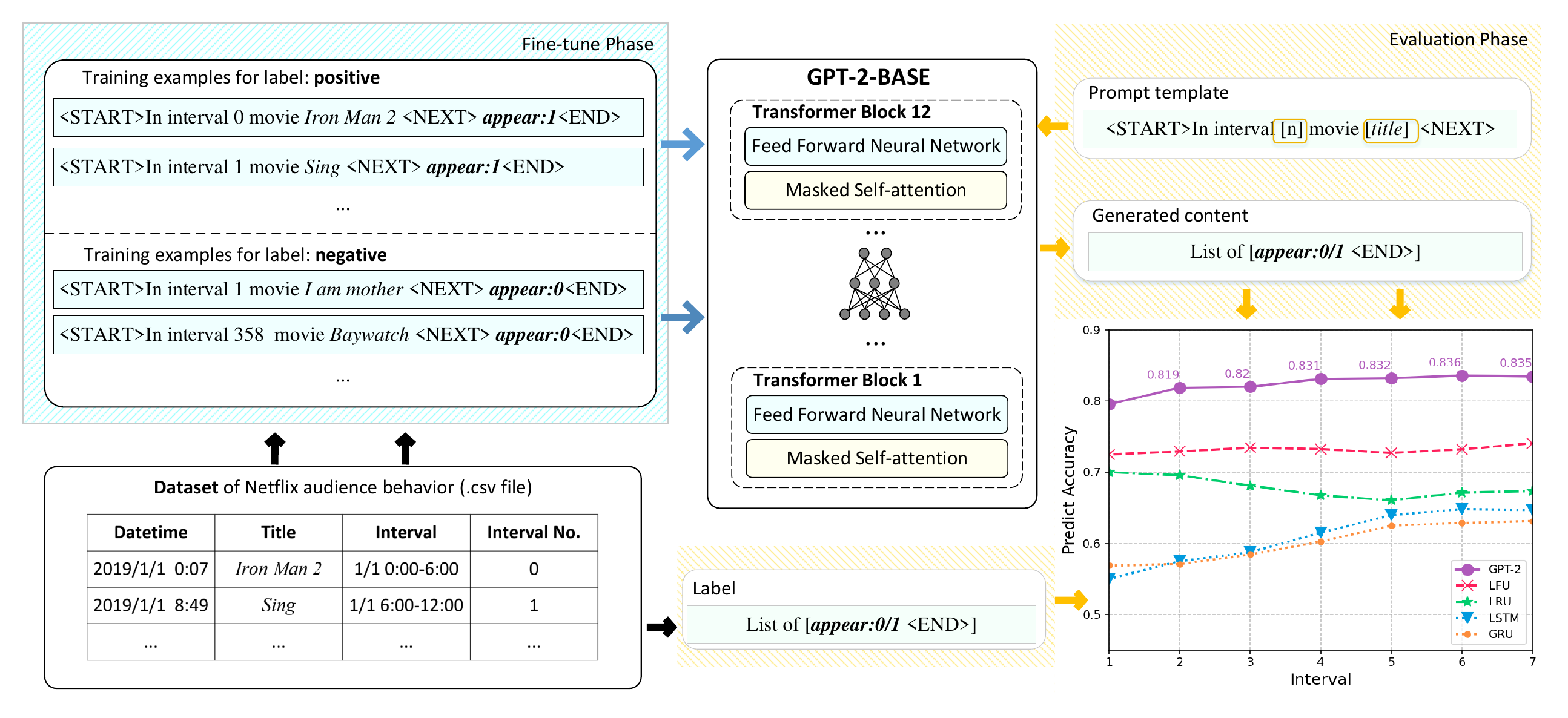}
	\caption{Edge LLM for popularity prediction: From data-sample template, fine-tuning to prediction accuracy.}
	\label{fig:prediction}
\end{figure*}

\subsection{Popularity Prediction}
Popularity prediction could significantly contribute to improving networking efficiency by adapting the C\&C resources to predicted demands \cite{9978680}. Considering the underlying principles of DNN structure, GPT-2 promises the ability to interpret users' preferences from historical visiting records from affiliated terminals at the RAN. Furthermore, by incorporating location-specific data, the edge LLM can be rather different to better capture personalized characteristics unique to each area. 

In order to test the prediction capability of the edge LLM (i.e., the GPT-2-base model), we take the Netflix audience behavior dataset as a showcase. In order to mitigate the data sparsity, the range of time is first divided into intervals based on a $6$-hour cycle and tagged a number. Subsequently, $20$ movies with the highest frequency are selected and labeled according to the presence or absence of each movie in a particular interval. Later, benefiting from the data format capability in CmP, the related historical information is converted into some natural languages conforming to a specific template. For example, ``\textit{In interval 1, movie `Iron man 2' appear :1}'' indicates the movie ``Iron man 2'' appears in the Interval $1$, which corresponds to some specific date-time given in the left-bottom part of Fig. \ref{fig:prediction}. Meanwhile, special tokens are added to create a prompt template that aids the language model in information comprehension and response generation. After direct fine-tuning, the edge LLM could generate labels following the prompt template format, i.e., whether the movie appears under the interval. Furthermore, to enhance model universality, we specifically utilize data from the last half year in the dataset for experimentation, and divide the dataset as the training set and test set according to the proportion $95\%$ to $5\%$. Fig. \ref{fig:prediction} finally presents the prediction accuracy of the edge LLM. It can be observed that GPT-2 exhibits an acceptable level of accuracy on this task, and significantly outperforms other classical algorithms (e.g., LSTM, GRU). Solely contingent on the edge LLM (i.e., the GPT-2-base model), this prediction capability demonstrates the potential of edge LLMs in interpreting and utilizing data within \texttt{NetGPT}.

\subsection{Intent Inference}

\begin{figure}[t]
	\centering
	\includegraphics[width=0.495\textwidth]{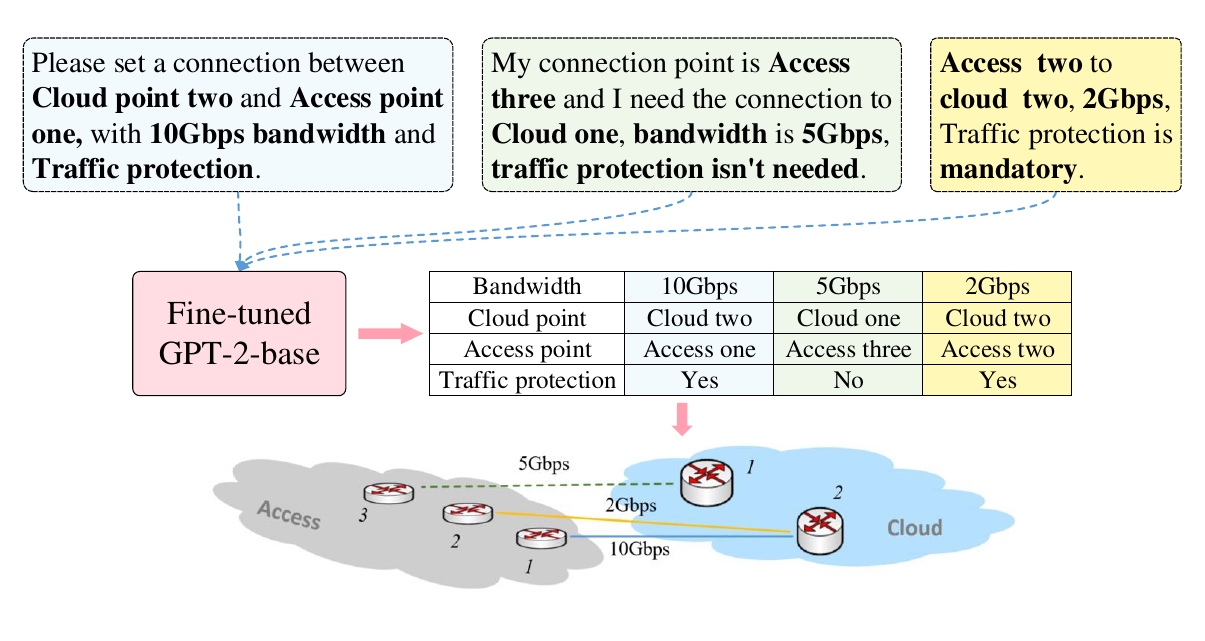}
	\caption{Edge LLM for intent inference.}
	\label{fig:intent}
\end{figure}
Intent-based networking aims to tackle the increased difficulty of applying template-based services to vertical business, and it needs to perceive the real-time requirements of customers before replacing the manual processes of configuring networks and reacting to network issues \cite{8968429}. In that regard, how to precisely understand the intent of customers and translate it into feasible network configuration belongs to one of the most fundamental issues. 

Coincidentally, edge LLMs exactly satisfy such intent-recognition process \cite{8968429} and benefit the accurate understanding of more verbal statements. In particular, by adopting the self-instruct approach~\cite{selfinstruct} as before, we first obtain a dataset encompassing $4,000$ input-keyword pairs, which map between linguistic network intents and typical network configuration keywords (e.g., bps and network protection). After fine-tuning on the dataset, 
there emerges the capability in the edge LLMs to understand and extract keywords from arbitrary natural language input. For example, the post-fine-tuning results in Fig. \ref{fig:intent} demonstrate that if one user wants to establish a $10$ Gbps connection from Access $1$ to Cloud $2$ with traffic protection, accurate keywords could be conveniently extracted by GPT-2-base model regardless of statement distinctions. Therefore, it avoids the cumbersome template design and customer learning process. In other words, compared to conventional NLP tools, LLMs manifest stronger capability towards fulfilling intent-driven networking, not only in understanding the semantics of user requests but also in the pragmatic application of these requests to network configuration tasks. Moreover, such a qualification for the intent-based network management also verifies the potential for LLMs to be reconfigured on-the-fly to accommodate speech patterns or evolving network commands, without the need for extensive re-training or manual intervention.

\section{Conclusion \& Future Work}
\label{sec:conclusion}
In this article, based on LLMs, we have advocated an AI-native network architecture, namely \texttt{NetGPT}, for provisioning network services beyond personalized generative content. In particular, through the effective cloud-edge LLM synergy, we have demonstrated the feasibility of \texttt{NetGPT} for location-based personalized services by deploying some representative open-source LLMs (e.g., GPT-2-base model and LLaMA model) at the edge and the cloud, and evaluating their coherent performance with the adoption of low-rank adaptation-based parameter-efficient fine-tuning techniques. Besides, we have comprehensively demonstrated the superiority of \texttt{NetGPT} over alternative cloud-edge or cloud-only techniques. On top of that, we have highlighted some substantial architectural changes (e.g., deep C\&C integration and a logical AI workflow) that \texttt{NetGPT} will require. As a by-product, we have presented a possible unified AI solution for network management \& orchestration empowered by edge LLMs through exemplifying the performance for popularity prediction and intent inference. 

While \texttt{NetGPT} is a promising AI-native network architecture for provisioning beyond personalized generative services, in this article, we have not discussed all of the major research challenges. For successful deployment of \texttt{NetGPT}, the following questions will need to be answered.
\begin{itemize}
    \item Given the success of LLaMA to shrink model sizes through effective algorithmic and structural updates, how to implement the inference and fine-tuning at the terminals, so as to satisfy the limited computing capability in cost-limited devices?
    \item Considering the continual evolution of knowledge, how to emulate new Bing\footnote{New Bing refers to GPT-empowered search engine available at \url{https://www.bing.com/new}.} and implement online learning-based LLMs to adapt to the dynamicity of wireless environment at the edge? Meanwhile, how to collect, distribute and process the data while maintain the essential privacy at the edge and cloud?
    \item Due to the limited sensitivity for numerical inference and possible deception effects, how to further improve the rigorousness of LLMs and what lessons can be learned from the latest LLM? Meanwhile, how to incorporate the evaluation metric of LLM to derive a suitable logical AI workflow? \item How to evaluate the interpretability and reliability of the model's outputs? How to bypass the hallucination effect of LLMs to meet stringent requirements for low-latency and ultra-reliability? More importantly, in addition to reinforcement learning from human feedback (RLHF), what techniques can be leveraged to satisfy users' stringent requirements amidst significant model uncertainties, before practically deploying LLMs in scenarios where decision-making critically relies on AI insights?
    
    \item How to incorporate the latest LMMs more effectively to provision personalized services beyond text? For different data modalities, how to effectively orchestrate heterogeneous computing resources to maximize operational efficiency?
    \item Regarding the cloud-edge collaboration, how to accurately learn the variations of availability of computing resources and develop adaptive resource management strategies to dynamically respond to fluctuating computational demands? Besides personalized assistance and recommendation systems, other real-world applications (e.g., AI copilot, embodied AI agent, etc.) can be explored to fully unleash the capability of \texttt{NetGPT}. Correspondingly, how can \texttt{NetGPT} be evolved to meet the emerging demands?
\end{itemize}


\section*{Author Biographies}
\textbf{Yuxuan Chen} is a PhD Candidate in Zhejiang University, Hangzhou, China. His research interests currently focus on Large Language Model in communication.

\textbf{Rongpeng Li} is an associate professor in Zhejiang University, Hangzhou, China. His research interests currently focus on networked intelligence for communications evolving.

\textbf{Zhifeng Zhao} is the Chief Engineer with Zhejiang Lab, Hangzhou, China. His research area includes collective intelligence and software-defined networks.

\textbf{Chenghui Peng} is a Principal Researcher of Huawei Technologies. His current research interests focus on 6G native AI architecture design.

\textbf{Jianjun Wu} is the Chief Researcher and Director of Future Network Lab, Huawei Technologies. He is leading the future network architecture design in Huawei.

\textbf{Ekram Hossain} is a Professor and the Associate Head (Graduate Studies) in the Department of Electrical and Computer Engineering at University of Manitoba, Canada.

\textbf{Honggang Zhang} is a principal researcher with Zhejiang Lab, Hangzhou, China. He is currently involved in research on cognitive green communications.
\end{document}